\documentclass[sigconf]{acmart}

\copyrightyear{2018}
\acmYear{2018}
\setcopyright{acmlicensed}
\acmConference[CIKM '18]{The 27th ACM International Conference on Information and Knowledge Management}{October 22--26,2018}{Torino, Italy}
\acmPrice{15.00}
\acmDOI{10.1145/3269206.3271683} \acmISBN{978-1-4503-6014-2/18/10}

\usepackage{booktabs} 
\usepackage{algorithm}
\usepackage{algorithmic}
\usepackage{amssymb}
\usepackage{latexsym,bm,amsmath,amssymb}
\usepackage{amsmath}
\usepackage{array}
\usepackage[normalem]{ulem}
\usepackage{footmisc}
\usepackage{graphicx}
\usepackage{multirow}
\usepackage{listings}
\usepackage[np, autolanguage]{numprint}
\usepackage{paralist}
\usepackage{enumitem}
\usepackage{mathtools}
\usepackage{soul}
\usepackage{epstopdf}
\usepackage{setspace}
\usepackage{pifont}
\usepackage[skip=0pt]{caption}
\usepackage{subfigure}
\usepackage{xcolor}
\usepackage{mathrsfs}

\newcount\colveccount
\newcommand*\colvec[1]{
        \global\colveccount#1
        \begin{matrix}
        \colvecnext
}
\def\colvecnext#1{
        #1
        \global\advance\colveccount-1
        \ifnum\colveccount>0
                \\
                \expandafter\colvecnext
        \else
                \end{matrix}
        \fi
}

\newcommand{\argmax}{\mathop{\mathrm{arg}\,\max}}

\newcommand{\paratitle}[1]{\vspace{1.5ex}\noindent\textbf{#1}}
\begin{document}

\title{Explicit State Tracking with Semi-Supervision\\for Neural Dialogue Generation}

\author{Xisen Jin\raisebox{3pt}{$\S$}\footnotemark[1], Wenqiang Lei\raisebox{3pt}{$\dagger$}\footnotemark[1], Zhaochun Ren\raisebox{3pt}{$\ddagger$}, Hongshen Chen\raisebox{3pt}{$\ddagger$}, Shangsong Liang\raisebox{3pt}{$\ast$},\\ Yihong  Zhao\raisebox{2pt}{$\ddagger$}, Dawei Yin\raisebox{2pt}{$\ddagger$}}
\affiliation{%
  \institution{
  \raisebox{1pt}{$\ddagger$}JD.com, Beijing, China\\
  \raisebox{1pt}{$\S$}Fudan University, Shanghai, China\\ 
  \raisebox{1pt}{$\dagger$}National University of Singapore, Singapore\\
  \raisebox{1pt}{$\ast$}King Abdullah University of Science and Technology, Thuwal, Saudi Arabia
  }
}
\email{xisenjin@gmail.com, wenqiang@comp.nus.edu.sg, renzhaochun@jd.com}
\email{chenhongshen@jd.com,shangsong.liang@kaust.edu.sa, ericzhao@jd.com, yindawei@acm.org}

\newcommand\blfootnote[1]
{%
	\begingroup 
	\renewcommand\thefootnote{}\footnote{#1}%
	\addtocounter{footnote}{-1}%
	\endgroup 
}

\begin{abstract}

The task of dialogue generation aims to automatically provide responses given previous utterances. Tracking dialogue states is an important ingredient in dialogue generation for estimating users' intention. However, the \emph{expensive nature of state labeling} and the \emph{weak interpretability} make the dialogue state tracking a challenging problem for both task-oriented and non-task-oriented dialogue generation:
For generating responses in task-oriented dialogues, state tracking is usually learned from manually annotated corpora, where the human annotation is expensive for training; for generating responses in non-task-oriented dialogues, most of existing work neglects the explicit state tracking due to the unlimited number of dialogue states.

In this paper, we propose the \emph{semi-supervised explicit dialogue state tracker} (SEDST) for neural dialogue generation. To this end, our approach has two core ingredients: \emph{CopyFlowNet} and \emph{posterior regularization}. Specifically, we propose an encoder-decoder architecture, named \emph{CopyFlowNet}, to represent an explicit dialogue state with a probabilistic distribution over the vocabulary space. To optimize the training procedure, we apply a posterior regularization strategy to integrate indirect supervision.
Extensive experiments conducted on both task-oriented and non-task-oriented dialogue corpora demonstrate the effectiveness of our proposed model. Moreover, we find that our proposed semi-supervised dialogue state tracker achieves a comparable performance as state-of-the-art supervised learning baselines in state tracking procedure.

\end{abstract}

\keywords{Dialogue generation, Dialogue state tracking, Semi-supervised learning, Posterior regularization}
\renewcommand{\shortauthors}{Xisen Jin, Wenqiang Lei, Zhaochun Ren, et al.}
\renewcommand{\shorttitle}{Explicit State Tracking with Semi-Supervision for Neural Dialogue Generation}

\maketitle
\blfootnote{* Work performed during an internship at JD.com.}
\vspace*{-1.5\baselineskip}

\section{Introduction}

In recent years, dialogue systems have received increasing attention in numerous web applications~\cite{Young2013POMDP,Ritter2011Data,Banchs2013IRIS,Ameixa2014Luke}. Existing dialogue systems can fall into two categories: non-task-oriented dialogue systems and task-oriented dialogue systems. Non-task-oriented dialogue systems aim to generate fluent and engaging responses, whereas task-oriented dialogue systems need to complete a specific task, e.g., restaurant reservation, along with a response generation process.
Employing neural networks to generate natural and sound responses, the task of \emph{neural dialogue generation} is playing an important role in dialogue systems~\cite{shang,vinyals2015neural,sordoni2015,li2016a,li2016b,serban2016building,Bordes2016Learning,wen-EtAl:2017:EACLlong,pei18emnlp}. 
In a dialogue, a \emph{dialogue state} refers to a full and temporal representation of each participant's intention~\cite{goddeau1996form}. 
Thus in neural dialogue generation, dynamically tracking dialogue states is the key for generating coherent and context-sensitive responses. 

Numerous dialog state tracking mechanisms with a limited state space have been proposed for task-oriented dialogue systems, e.g., hand-crafted rules~\cite{goddeau1996form,wang2013simple}, conditional random fields~\cite{lee2013recipe,lee2013structured,ren2013dialog}, maximum entropy~\cite{williams2013multi}, and neural networks~\cite{henderson2013deep}. 
As a state-of-the-art work, \emph{explicit dialog state tracking via an interpretable text span}  has been preliminarily attempted on task-oriented dialogue systems  ~\cite{sequicity}. Differently, in non-task-oriented dialogue systems, most of existing state tracking approaches employ a fixed-size latent vector to represent the whole dialogue history~\cite{serban2016building}. Though these solutions are capable for chit-chat conversations, they fail to distinguish similar concepts or entities(e.g., product names) which are often key information in technical and transactional  domains~\cite{Bordes2016Learning}. Moreover, these latent vectors have weak interpretability. However, existing solutions to explicit state tracking cannot be applied in non-task-oriented dialogue systems, since these supervised approaches typically require large amounts of manually annotated dialogue states.

Accordingly, the neural dialogue generation faces a dilemma between reducing the expense of data annotation and improving the performance of dialogue state tracking: (1) For both task-oriented and non-task-oriented dialogue generation, existing explicit approaches requires a large amounts of manually labeled data to train the state tracker~\cite{sequicity}. Heavily relying on the expensive annotated corpus, these methods lead the state tracker extremely difficult to be transferred to new scenarios or extended to a larger state space. 
(2) Most of unexplainable state trackers in non-task-oriented dialogue generation are not capable of explicitly tracking long-term dialogue states, limiting their capability in complicated domains. 
To tackle the above challenges, our focus is on developing methods to construct an explicit dialogue state tracker with unlabeled data for neural dialogue generation.

In this paper, we propose a semi-supervised neural network, named the \emph{semi-supervised explicit dialogue state tracker} (\textbf{SEDST} for short), to explicitly track dialogue states for both task-oriented and non-task oriented dialogue generation with a text span. 
Along with SEDST, we propose a novel encoder-decoder architecture based on copying mechanism~\cite{gu2016incorporating}, called \emph{CopyFlowNet}, to represent dialogue states with explicit word sequences. We infer these word sequences, i.e., text spans, through a probabilistic distribution over the vocabulary space.
To optimize the training procedure of SEDST, we employ a posterior regularization strategy to integrate indirect supervision from unlabeled data. Thus SEDST is compatible for both supervised and unsupervised learning scenarios. 

In our experiments, we verify the effectiveness of our proposed method in both task-oriented dialogue generation and non-task-oriented dialog generation, respectively. We find that SEDST, under 50\% data annotated setup, outperforms state-of-the-art task-oriented dialogue generation baselines, as well as outperforms non-task-oriented baselines under no data annotated setup. Moreover, we deeply analyze and verify the effectiveness of the posterior regularization strategy in incorporating indirect supervision for the dialogue state tracking.

\smallskip\noindent
\noindent To sum up, our main contributions can be summarized as follows:
\begin{itemize}
\item We focus on tracking explicit dialogue states with semi-supervision for neural dialogue generation.
\item We propose a semi-supervised neural dialogue generation framework, called \textbf{SEDST}, for both task-oriented and non-task-oriented dialogue systems. 
\item We propose an explicit dialogue state tracker, CopyFlowNet, with implicit copyNets and posterior regularization. 
\item We verify the effectiveness of SEDST in our extensive experiments on both task-oriented copora and non-task-oriented copora.
\item We deeply study and analyze the performance of SEDST and other widely used two-stage decoding models in dialogue systems.
\end{itemize}

\noindent We introduce related work in \S\ref{section2}. We provide preliminaries in \S\ref{section3} and describe our approach in \S\ref{section4}. Then, \S\ref{section5} details our experimental setup, \S\ref{section6} presents the results, and \S\ref{section7} concludes the paper.

\section{Related Work}
\label{section2}

We detail our related work on two lines: neural dialogue generation and dialogue state tracking.

\subsection{Neural dialogue generation}

Neural dialogue generation aims at generating natural-sounding replies automatically to exchange information, e.g., knowledge~\cite{Young2013POMDP,shawar2007,chen18www}.
As a core component of both task-oriented and non-task-oriented dialogue systems, neural dialogue generation has received a lot attention in recent years~\cite{Young2013POMDP,Ritter2011Data,Banchs2013IRIS,Ameixa2014Luke}. Among all these approaches, sequence-to-sequence structure neural generation models~\cite{shang,vinyals2015neural,sordoni2015,li2016a,li2016b,serban2016building,cao2017,chen18www} have been proved to be capable in multiple dialogue systems with promising performance. 
Several approaches have been proposed to softly model language patterns such as word alignment and repeating into sequence-to-sequence structure~\cite{bahdanau2015neural,xing2016topic,gu2016incorporating,serban2017multiresolution,cao2017}.
\citet{xing2016topic} employ attention mechanism~\cite{bahdanau2015neural} to dynamically incorporates contextual information into response generation. 
\citet{gu2016incorporating} propose a copy mechanism to consider additional copying probabilities for contextual words in forum conversations. 
\citet{serban2017multiresolution} decodes coarse tokens before generating the fulls response.
\citet{cao2017} tackle the boring output issue of deterministic dialogue models by introducing a latent variable model for one-shot dialogue response.
In \cite{dawnet}, the authors selects and predicts explicits keywords using  before response generation. 
\citet{zhang2017acl} use a log-linear model to represent the desired distribution and inject the prior knowledge by a posterior regularization.
Additionally, recent work verify that reinforcement learning is a promising paradigm when state and action spaces are carefully designed~\cite{williams-asadi-zweig:2017:Long,dhingra-EtAl:2017:Long1}. 

\subsection{Dialogue state tracking}
Dialogue state tracking is an important ingredient of the dialogue generation.
Traditional methods utilize hand-crafted rules to select the dialogue state~\cite{goddeau1996form}. Relying on the most likely results from an natural language understanding (NLU) module~\cite{Perez17}, these rule-based systems hardly models uncertainty, which is prone to frequent errors~\cite{williams2014web,Perez17}. \citet{Young2010The} propose a distributional dialogue state for statistical dialog system and maintain a distribution over multiple hypotheses facing with noisy conditions and ambiguity. Another typical form of dialogue state is in the form of a probability distribution over each slot for each turn~\cite{Williams2012A,williams2013dialog}.  

In task-oriented dialogue systems, end-to-end neural networks have been successfully employed for tracking dialogue states via interacting with an external knowledge base~\cite{wen-EtAl:2017:EACLlong,eric2017key,Bordes2016Learning,williams-asadi-zweig:2017:Long}.
\citet{wen-EtAl:2017:EACLlong} divide the training procedure into two phases: the dialogue state tracker training, and the whole model training. 
\citet{mrkvsic2016neural} proposed a dialogue state tracker based on word embedding similarities.
\citet{eric2017key} implicitly model a dialogue state through an attention-based retrieval mechanism to reason over a key-value representation of the underlying knowledge base. 
  \citet{Bordes2016Learning} memories the dialogue context in a memory module and repeatedly queries and reasons
about this context to select an adequate system response.
  Instead of employing symbolic knowledge queries, \citet{dhingra-EtAl:2017:Long1} propose an induced ``soft'' posterior distribution over the knowledge base to search matching entities. 
\citet{sequicity} proposed an extendable framework to track dialogue states with a text span including the constraints for a knowledge base query.

In non-task-oriented dialogue systems such as forum conversations, lots of efforts have been made to keep track of the dialogue process in multi-turn settings~\cite{serban2016building,Sordoni2015hre,hvred}. \citet{serban2016building} and \citet{Sordoni2015hre} incorporates hierarchical structures in word and sentence levels to encourage cohesive multi-turn dialogue generation. \citet{hvred} utilize a latent variable at the sub-sequence level in a hierarchical setting. \citet{chen18www} add a hierarchical structure and a variational memory module into a neural encoder-decoder network. However, all these latent vectors and latent memories are unexplainable, which makes it challenging to verify the effectiveness of dialogue state tracking. 
Moreover, these unexplainable latent vectors fail to distinguish  distinctive concepts with similar vector representation, e.g., product names. 

\smallskip\noindent
Our work differs from previous work in the following important ways: (1) We represent dialogue states in text spans explicitly; (2) We propose the CopyFlowNet which enables semi-supervised and unsupervised training of the state tracker; (3) We propose a novel training method incorporating posterior regularization to improve the robustness.

\begin{table}[!t]
\caption{Glossary.}
  \centering
  \begin{tabular}{ll}
  \hline
  \textbf{Symbol} & \textbf{Description}\\
  \hline
  $D$  & a dialogue session  \\
  $U$  & a user utterance  \\
  $R$  & a machine response \\
  $S$  & a text span for dialogue state tracking\\
  $X$  & an input sequence\\
  $Y$  & an output  sequence\\
  $s$  & an element in a text span \\
  $x$  & an element in a source sequence \\
  $y$  & an element in a target sequence \\
  $N$  & length of an utterance, text span and a machine response\\
  $w_x$  & a word input in source sequence \\
  $\mathbf{h}$ & a hidden vector generated in a GRU.\\
  $e$ & an entity obtained in the knowledge base search\\
  $\mathcal{I}$ & a user's intention in a dialogue\\
  $\mathcal{V}$  & vocabulary \\
  $\mathcal{K}$  & a knowledge base \\
  $\mathcal{A}$ & a set of annotated training instances \\
  $\mathcal{U}$ & a set of unannotated training instances \\
  \hline
  \end{tabular}
  \label{tab:glossary}
  \vspace{-1.0em}
\end{table}

\section{Preliminaries}
\label{section3}

In this section, we first formalize task-oriented and non-task-oriented dialogue systems. Thereafter, we introduce the approach of tracking dialogue state using text spans.

\subsection{Problem formulation}

Before introducing our method for dialogue state tracking, we introduce our notation and key concepts.
Table~\ref{tab:glossary} lists our notations in this paper. 
Given $T$ dialogue turns, a \emph{dialogue session} $D$ consists of a sequence of utterances, i.e., $D=\{U_{1}, R_1, U_2, R_2 .., U_{T}, R_T\}$, where $U$, $R$ refers to responses from a user and a machine respectively. At the $t$-th turn, given the current user utterance $U_t$ and  historical records $U_{1}, R_1, U_2, R_2..,U_{t-1},R_{t-1}$, the dialogue system generates a response $R_t$. Probabilistically, the system generates $R_t$ via maximizing the probability $P(R_t|U_{1}, R_1, U_2, R_2..,U_{t-1},R_{t-1})$, shown in Eq.~\ref{eq:rt1}:

\begin{equation}
\label{eq:rt1}
R_t = \argmax P(R_t|U_{1}, R_1, U_2, R_2..,U_{t-1},R_{t-1}, U_t),
\end{equation}

\noindent \emph{Task-oriented dialogue system} includes the \emph{task completion} component which is specified by users (e.g., reserving an restaurant)~\cite{wen2017network}. To further formulate the task completion, we denote the user intention as $\mathcal{I}$. In this paper, we simplify the user's intention as a specific entity $e$, searched in knowledge bases at the end of a dialogue.

\noindent \emph{Non-task oriented systems}, differently, just focus on generating engaging and coherent responses. Typically they involve broader domains and more complicated contextual information~\cite{serban2017multiresolution,dawnet}. Note that it is difficult to find a clear boundary between task oriented dialogue systems and non-task oriented dialogue systems. To clarify the difference between these two types of dialogue systems, in this paper 
any dialogue system including knowledge base interactions refers to a task-oriented dialogue system.

Dialogue state tracking is the key component for both non-task-oriented dialogue systems and task oriented dialogue systems. 
In non-task oriented dialogue systems, dialogue state tracking is a key to generate context-aware and coherent responses. 
Whereas in task-oriented dialogue systems, dialogue state tracking is becoming mandatory since it has to capture users' request and constraints for knowledge base search. 

\subsection{Dialogue state tracking using text spans}

Text-span based dialogue state trackers have been proposed to manifest  simplicity along with better interpretability~\cite{wen2017network,sequicity}. 
At $t$-th turn, such state tracker employs a text span $S_t$ (i.e., a sequence of words)  to track dialogue states. $S_t$ aims at summarizing past utterances and responses (i.e., $U_1$, $R_1$, $U_2$, $R_2$,...,$U_{t-1}$,$R_{t-1}$$,U_t$). 

We follow the notation from the Sequicity~\cite{sequicity}, which is a state-of-the-art dialogue state tracker for the task-oriented dialogue generation. 
Sequicity defines a text span over the full vocabulary space called \emph{bspan} (denoted as $S_t$), which records all ``requestable slots'' and ``informable slots''~\cite{wen2017network}, each separated with delimiters. Informable slots track the constraints which are used for knowledge base search while requestable slots record what users are looking for in current dialogues. 
With bspan, knowledge base search can be performed by taking informable slots as search constraints. As such, task completion can be converted as a problem of generating a text span $S_t$ at each turn.

Generating a text span $S_t$ of keywords also improves performance of single response generation~\cite{serban2017multiresolution,dawnet}. We elaborate this strategy into dialogue state tracking within multi-turn dialogues for non-task oriented dialogues. In our work, we define a \emph{state span} as a text span that indicates a dialogue state.
Shown in Eq.~\ref{eq1:st}, in this paper the problem of both task-oriented and non-task-oriented dialogue generation can be decomposed into two successive steps: (1) generating a state span $S_t$; (2) generating the response $R_t$. 

\begin{flalign}
\label{eq1:st}
\begin{split}
& S_t = \argmax P(S_t|U_{1}, R_1, U_2, R_2..,U_{t-1},R_{t-1}, U_t), \\
& R_t = \argmax P(R_t|S_t, U_{1}, R_1, U_2, R_2..,U_{t-1},R_{t-1}, U_t).
\end{split}
\end{flalign}

\section{Method}
\label{section4}

\begin{figure*}[!t]
  \centering 
\includegraphics[width=0.9\textwidth,height=0.4\textwidth]{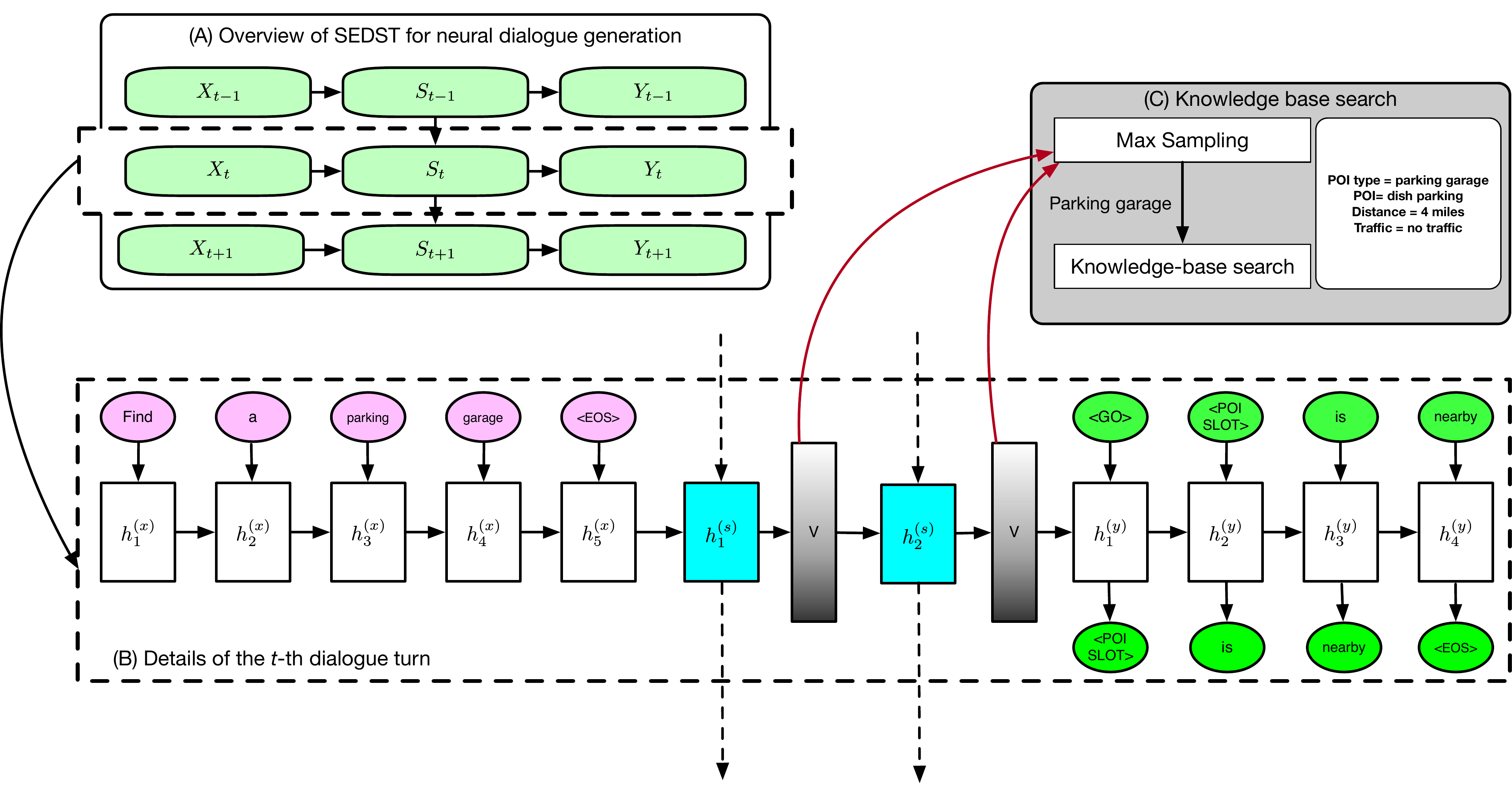} 
  \caption{Overview of our dialogue model with semi-supervised explicit dialogue state tracker. (A) provides an overview of dialogue response generation process. Each arrow represents a \emph{copyflow}(see ~\ref{subsec:copynet}). (B) provides details of the t-th dialogue turn. 
  (C) represents the knowledge base interaction for task-oriented dialogues. ``Parking garage'' is the constraint for knowledge base search in the example. Ellipses are the input and output tokens and rectangles denote neural hidden vectors.} 
  \label{fig:overview}
\end{figure*}

In this section, we propose our semi-supervised explicit dialogue state tracker, abbreviated as \textbf{SEDST}. We start by providing an overview of SEDST. We then describe the CopyFlowNet architecture and detail our posterior regularization for optimizing the model training.

\subsection{Overview}

We propose the \emph{semi-supervised explicit state tracker} (\textbf{SEDST}) to track dialogue states with explicit text spans under no or few data annotation setup. SEDST includes two main ingredients: (1) CopyFlowNet; (2) posterior regularization. Along with SEDST, we propose an encoder-decoder architecture, \emph{CopyFlowNet}, to generate explicit state spans by copying from existing sequences on top of normal generation process. Finally, we provide an optimized training procedure of SEDST, where we apply a posterior regularization strategy to improve the robustness of our model.

Figure~\ref{fig:overview} provides an overview for state tracking and response generation process in SEDST. The turn-level overview presents how the text span is utilized for state tracking for multi-turn dialogues, where each arrow represents a \emph{copyflow}(see \ref{subsec:copynet}). The sentence level illustration presents encoding and decoding procedure. We also illustrate how the knowledge-base interaction is performed with state spans for task-oriented dialogues, where we follow~\cite{sequicity}.

\subsection{CopyFlowNet}
\label{subsec:copynet}

In this section we detail the \emph{CopyFlowNet} architecture.
At $t$-th dialogue turn, for simplicity we set the length of $U_t$ and $R_{t-1}$ to $N$; CopyFlowNet first encodes the concatenation of previous response $R_{t-1}$ and current user utterance $U_t$ (i.e., $R_{t-1}U_t=w_{x_1},w_{x_2},...,w_{x_{2N}}$) 
with gated recurrent unit (GRU) encoders~\cite{chung2014empirical}:

\begin{equation}
\label{encoder}
\mathbf{h}^{(x)}_1,\mathbf{h}^{(x)}_2,...,\mathbf{h}^{(x)}_{2N}=GRU(\mathbf{w}_{x_1},\mathbf{w}_{x_2},...,\mathbf{w}_{x_{2N}}),
\end{equation}
 
\noindent At $t$-th turn, each token in $S_t$ and $R_t$ is decoded with both probability of direct generation and that of copying from a previous sequence. 
 
We employ an attention decoder~\cite{bahdanau2015neural} to calculate the generation probability of $S_t$ and $R_t$ individually. 
When decoding an output $Y \in \left\{ {S_t,R_t} \right\}$ at $t$-th turn, the decoder attends back to hidden states $\mathbf{h}^{(x)}$ of the input $X$ to compute attention vectors.
We calculate attention vectors with $R_{t-1}U_t$ when generating $S_t$, and concatenate $S_t$ to $R_{t-1}U_t$ when generating $R_t$. For the generation of the $j$-th word in $Y$, $y_j$, we calculate its attention score $a_{ij}$ as follows~\footnote{We also tried dot-product attention and got similar experimental results.}: 
\vspace*{-.4\baselineskip}
\begin{equation}
\label{attn1}  a_{ij}=softmax(\mathbf{v_1}^Ttanh(\mathbf{W}_1{\mathbf{h}^{(x)}_i}+\mathbf{W}_2{\mathbf{h}^{(y)}_{j-1}}))
\end{equation}

\noindent where ${\mathbf{h}^{(x)}_i}$ and ${\mathbf{h}^{(y)}_{j-1}}$ indicate a hidden state of $i$-th word in $X$ and a hidden state of $j-1$-th word in $Y$, respectively.
$\mathbf{v}_1$, $\mathbf{W}_1$ and $\mathbf{W}_2$ are learnable parameters.
Given $a_{ij}$, the decoder then generates a hidden representation $\mathbf{h}^{(y)}_j$ of the $j$-th word using GRU:

\begin{flalign}
\label{attnsum}
\begin{split}
& {\tilde{\mathbf{h}}^{(x)}}_j=\sum_{i} a_{ij}{\mathbf{h}^{(x)}_i}, \\
& h_j^{(y)}=GRU(w_{y_{j-1}}, h_{j-1}^{(y)},\tilde{\mathbf{h}}^{(x)}),
\end{split}
\end{flalign}

\noindent where $\tilde{\mathbf{h}}^{(x)}$ refers to an attention vector by summing up all ${\mathbf{h}^{(x)}_i}$, and $w_{y_{j-1}}$ indicates the $j-1$ word in $Y$. Thereafter, we 
get the generation probability distribution $\mathbf{p}_j^g$ for the $j$-th output word:

\begin{equation}
\label{attnproj}
  \mathbf{p}_j^g=\frac{1}{Z}e^{({\mathbf{W_3}\mathbf{h}^{(y)}_j})},
\end{equation}

\noindent where $Z$ is a shared normalization term and $\mathbf{W_3}$ is a learnable parameter. $\mathbf{p}_j^g$ can be considered as the normalized projection of $\mathbf{h}^{(y)}_j$ at the output space.

We employ copyNets ~\cite{gu2016incorporating} to compute the probability of copying words from a deterministic input $X$ when decoding $y_j$. We denote the probability as $p^{c(X)}(y_j)$, and the aggregated probability distribution as $\mathbf{p}_j^{c(X)}$. This probability of generating an word output $y_j$ by copying from $X$ is calculated as:

\begin{equation}
p^{c(X)}(y_j)=
\begin{cases}
	\frac{1}{Z} \sum_{i:w_{x_i}=y_j} e^{\psi(w_{x_i})} , y_j \in X \,  \\
    0, otherwise
\end{cases}
\end{equation}

\noindent where $Z$ is a shared normalization term. $\psi(x_i)$ is the weight of copying the $i$-th word in $X$, calculated in Eq.~\ref{eq:psi}:

\begin{equation}
\label{eq:psi}
\psi(y_j=w_{x_i})=\mathbf{v_2}^T tanh(\mathbf{W_4}{\mathbf{h}^{(x)}_i}+\mathbf{W_5}{\mathbf{h}^{(y)}_j}),
\end{equation}

\noindent where $\mathbf{v}_2$, $\mathbf{W_4}$ and $\mathbf{W_5}$ are learnable parameters.

Since $S_{t}$ could be unlabeled, we may have to copy from a nondeterministic model-generated distribution. 
Also to enable multiple-step copying along dialogue turns, we propose \emph{implicit copyNets} to copy words with high confidence from a sequence of nondeterministic word distribution. The copying probability is calculated as:

\begin{equation}
p^{c(X)}(y_j) =\frac{1}{Z} \sum_{i=1}^{|X|} p_i(w_{x_i}=y_j) e^{\psi(w_{x_i})},
\end{equation}

\noindent where $p_i(w_{x_i}=y_j)$ is the probability that the $i$-th word in the nondeterministic input $X$ equals $y_j$. Note that in cases where the source sequence is deterministic such as $U_{t}$, $p_i(w_{x_i}=y_j)$ degenerates to a binary indicator where $p_i(w_{x_i})=1$ if $w_{x_i} = y_j$. At such situation, implicit CopyNets are equivalent to original copyNets. 

Eventually, we demonstrate organization details of CopyFlowNet. A \emph{copy flow} from $X$ to $Y$ refers to a procedure involving copying probability from $X$ at the generation of $Y$. Specifically, at a single turn $t$ in our model, there are copy flows from $R_{t-1}U_t$ to $S_t$ and from $S_t$ to $R_t$; there is another copy flow from $S_{t-1}$ to $S_t$ in adjacent dialogue turns. The name of \emph{copyflows} is illuminated by the phenomenon that the information flow of $S_1,S_2,...,S_t$ relay along the dialogue turns by copying and finally attend to the generation of $R_t$. Formally, the probability distribution of the $j$-th word in $S_t$ and $R_t$ are calculated as:

\begin{flalign}
\label{copyflowformula}
\begin{split}
& \mathbf{p}_j(S_t) = 
\begin{cases}
\mathbf{p}_j^g(S_t) + \mathbf{p}_j^{c(S_{t-1})}(S_t) + \mathbf{p}_j^{c(R_{t-1}U_t)}(S_t) , t>0 \\
\mathbf{p}_j^g(S_t) + \mathbf{p}_j^{c(R_{t-1}U_t)}(S_t) , t=0
\end{cases} \\
& \mathbf{p}_j(R_t) = \mathbf{p}_j^g(R_t) + \mathbf{p}_j^{c(S_t)}(R_t).
\end{split}
\end{flalign}

Every {copy flow} has an intuitive explanation. {Copy flows} from $R_{t-1}U_t$ to the state span $S_t$ and from $S_t$ to the response $R_t$ enable the model to cache ``keywords'' in $S_t$, and then copy them again to the final response $R_t$. It is inspired from the observation that the keywords indicating dialogue states are prone to co-occur among user inputs and system responses. Given an example dialogue turn:

\begin{itemize}
\item User: I want to book a French restaurant.
\item Model: C\^{o}te Brasserie is a French restaurant with good reputation.
\end{itemize}

\noindent The word ``French'' is the key word between the utterances which should be copied from $U_t$ to $S_t$ and then be copied to $R_t$. The model can also generate new words from the full vocabulary with normal generation process at $S_t$ for further copying to $R_t$. Note that all these actions can be learned even in unsupervised settings. Besides, the {copy flow} from the previous state span $S_{t-1}$ to current state span $S_t$ encourages useful dialogue states to be passed through dialogue turns, which intrinsically provides a solution for long-term dependency, as well as enables copying-mechanism to learn information from co-occurrence of keywords across different dialogue turns.

\subsection{Posterior regularization}
\label{sectionPR}

Due to the scarcity of supervised signal in state span $S_t$, the training of the state span could be unstable, especially when $S_t$ is not annotated. To tackle this problem, we apply posterior regularization on $S_t$ to train the state tracker more stably.

The normal forward pass of the network in the previous discussion parameterized by $\Theta$ actually computes the prior probability distribution of $S_t$:

\begin{equation}
P_\Theta(S_t|R_{t-1},S_{t-1},U_t)=\Pi_i \mathbf{p}(s^{(i)}_t|s^{(<i)}_t,R_{t-1},S_{t-1},U_t),
\end{equation}

We then build another network, the posterior network, which learns the posterior distribution of $S_{t}$ with more informative inputs. It adopts the same structure but it is separately parameterized~\footnote{The response decoder is shared between two networks in semi-supervised learning senarios.} with $\Phi$. It takes $R_{t-1}$, $S_{t-1}$, $U_{t}$, and $R_{t}$ as input, and calculates the posterior distribution of $S_t$ as follows:

\begin{equation}
Q_\Phi(S_t|S_{t-1},R_{t-1},U_t,R_t) = \Pi_i \mathbf{q}(s^{(i)}_t|s^{(<i)}_t,R_{t-1},S_{t-1},U_t,R_t),
\end{equation}
where $R_{t}$ is concatenated to the input at the encoder. Note that only the prior network works during testing time, while the the posterior network only directs the prior network in training. 

The idea of posterior regularization is to force the prior distribution $P$ to approximate the posterior distribution $Q$ learned from more informative inputs. We utilize KL-divergence to regularize these two distributions. Given multinomial distributions $\mathbf{p}_i$ and $\mathbf{q}_i$ over the vocabulary space, the KL-divergence from $\mathbf{p}_i$ to $\mathbf{q}_i$ is calculated as follows:

\begin{equation}
\label{eq:klpq}
KL(\mathbf{q}_i||\mathbf{p}_i) = \sum^{|\mathcal{V}|}_l q^{(l)}_i log(\frac{q^{(l)}_i}{p^{(l)}_i}),
\end{equation}

The training process of the model varies according to the amount of labeled data. In supervised or semi-supervised learning, we maximize the joint log-likelihood~\cite{wen2017latent} for response and state span generation along with posterior regularization. The learning objective thus comprises three sub-objectives.

\begin{equation}
\begin{aligned}
& \mathcal{L}_1=-\sum^{\mathcal{A}\cup \mathcal{U}}log[P(R_t|R_{t-1}, U_t, S_t)] 
\\&- \sum^{\mathcal{A}} log[P_{\Theta}(S_t|R_{t-1},U_t,S_{t-1})Q_{\Phi}(S_t|R_{t-1},U_{t}, S_{t-1}, R_t)]
\\&+\lambda \sum^{\mathcal{U}} \sum^{N}_{i=1} KL(\mathbf{q}_i||\mathbf{p}_i)\, ,
\end{aligned}
\end{equation}
where $\mathcal{A}$ and $\mathcal{U}$ denote annotated and unannotated training instances respectively. $N$ is the length of the state span.

However, maximum likelihood estimation on the distribution $Q_{\Phi}(S_t|S_{t-1}, R_{t-1},U_{t}, R_t)$ can not be applied when $S_t$ is completely unannotated. We therefore indirectly train the posterior distribution $Q_\Phi(\hat{S}_t|S_{t-1},R_{t-1},U_t,R_t)$ with different generation objectives at the decoder of the posterior network. In detail, we feed $R_{t-1}$, $U_{t}$, $R_{t}$ at its encoder and train the model to reconstruct them at its decoder. The posterior network thereby learns to cache most informative words in both contexts and responses at $\hat{S}_t$ via a structure of auto-encoders. The prior distribution $P_\Theta(S_t|R_{t-1},S_{t-1},U_t)$ is regularized towards this posterior distribution with KL-divergence, and the learning objective can be written as
\vspace*{-.5\baselineskip}
\begin{equation}
\label{unsuploss}
\begin{aligned}
\mathcal{L}_2&=-\sum^{\mathcal{U}} log[P(R_t|R_{t-1}, U_t, S_t)] 
\\&- \sum^{\mathcal{U}} log[Q_{\Phi}(R_{t-1},U_{t},R_{t}| \hat{S}_t)]
\\&+\lambda \sum^{\mathcal{U}} \sum^{N}_{i=1} KL(\mathbf{q}_i||\mathbf{p}_i)\, .
\end{aligned}
\end{equation}

\noindent which can be interpreted as response generation loss, reconstruction loss and regularization loss respectively . 
Following~\cite{higgins2016beta}, we employ a factor $\lambda$ as a trade-off factor. See~\S\ref{subsec:exp:training} for more details.

\section{Experimental Setup}
\label{section5}

\subsection{Research questions}
We list the research questions that guide the remainder of the paper:
(1) \textbf{RQ1}: what is the overall performance of SEDST in task-oriented neural dialogue generation? (See \S\ref{sec61})
(2) \textbf{RQ2}: How much does unlabeled data help dialogue state tracking in task-oriented dialogues in our model? (See \S\ref{sec61}.) 
(2) \textbf{RQ3}: Is our explicit state tracker helpful in response generation in non-task-oriented dialogues? (See \S\ref{sec62})
(4) \textbf{RQ4}: Does posterior regularization improve the model performance? (See \S\ref{sec63})
(5) \textbf{RQ5}: Can our dialogue state tracker generate explainable and representative words? Can it tackle the long-term dependency in a dialogue generation? (See \S\ref{sec64})

\smallskip\noindent
\noindent Next, we introduce the datasets in \S \ref{subsec:exp:dataset}. The baselines are listed in \S \ref{subsec:exp:baselines} and evaluation methods are depicted in \S \ref{subsec:exp:eval}. Details of the training setting are described in \S \ref{subsec:exp:training}.

\subsection{Datasets}
\label{subsec:exp:dataset}
In order to answer our research questions, we work with two task-oriented dialogue corpora: {Cambridge Restaurant Corpus} and {Stanford In-Car Personal Assistant Corpus}; two non-task-oriented dialogue corpora: {Ubuntu Technical Corpus} and {JD.com Customer Service Corpus}. Details of our datasets are described as follows:

\paratitle{Cambridge Restaurant Corpus}
Cambridge Restaurant corpus is used to design a dialogue system to assist users to find a restaurant in the Cambridge, UK area ~\cite{wen-EtAl:2017:EACLlong}. Customers can use three informable slots (\textit{food, pricerange, area}) to constrain the search. 
This dataset contains $99$ restaurants, and $676$ clean dialogues out of $1500$ dialogue turns. There are $99$ possible informable slot values. We split the corpus by 3:1:1 as training, validation and test sets. 

\paratitle{Stanford In-Car Personal Assistant Corpus}
The Stanford driver and car assistant corpus is a multi-turn multi-domain task-oriented dialogue dataset\footnote{https://nlp.stanford.edu/blog/a-new-multi-turn-multi-domain-task-oriented-dialogue-dataset/}. This dataset includes three distinct domains: calendar scheduling, weather information retrieval, and point-of-interest navigation. There are two modes, namely \textit{Driver} and \textit{Car Assistant}. There are $284$ informable slot values for state tracking. Each dialogue is associated with a separate knowledge base (KB) with about $7$ entries. The corpus contains $2425$ ,$302$, $302$ dialogues for training, validation and testing.

\paratitle{Ubuntu Technical Corpus}
Ubuntu Dialogue Corpus \cite{ubuntu} is an English multi-turn dialogue corpus containing about $487337$ dialogues extracted from the Ubuntu Internet Relayed Chat channel. A conversation begins with an Ubuntu-related technical problem, and follows by the responses to the questions. The corpus consists of $448833$, $19584$, $18920$ dialogues of training, validation, testing, respectively. Though the corpus is domain specific, the task and slot-values are not explicitly specified. 

\paratitle{JD.com Customer Service Corpus}
JD.com customer service corpus~\cite{chen18www} is a large real-world dataset for online shopping after-sale service. The conversation is between a customer and a customer service staff. It contains $415,000$ dialogues for training, $1,5000$ dialogues for validation, and $5,005$ for testing. We exclude template dialogue turns that the staff or customers merely thanks with keyword filtering.

\begin{table*}[t]
\caption{RQ1\&RQ2\&RQ4: Performance on Cambridge Restaurant corpus. N/A indicates the model fails to produce valid result on this dataset. Best performance is marked bold for each supervision proportion}
\centering
\begin{tabular}{c|c|c|c|c}
\hline
            &\textbf{Supervision Proportion} & \textbf{BLEU} &\textbf{Joint Goal Accuracy} &\textbf{Entity Match Rate}\\
\hline \hline

\multirow{3}{*}{\emph{SEDST}} &0\%  &\textbf{0.201} &\textbf{0.684} &\textbf{0.649}\\
                                     &25\% &\textbf{0.225} &\textbf{0.867} &\textbf{0.858}\\
                                     &50\% &\textbf{0.236} &\textbf{0.945} &\textbf{0.927}\\
\hline
\multirow{3}{*}{\emph{SEDST$\backslash$PR}}  & 0\%	 &0.199 &0.679 &0.422\\
                                  & 25\% &0.213 &0.854 &0.848\\
								  & 50\% &0.192 &0.911  &0.901\\ 
\hline
\multirow{3}{*}{\emph{SEDST(without unlabeled data)}}	
                                    &0\%	&-  &- &-\\ 
									&25\%	&0.091 &0.827 &0.807 \\
                                    &50\%	&0.122 &0.896 &0.899\\                  
\hline
\emph{SEDST(fully supervised)} &100\% &\textbf{0.244}  &\textbf{0.962} &\textbf{0.955}\\    
\hline
\emph{NDM} & 100\% & 0.239 & 0.921 & 0.902 \\
\hline
\emph{Neural Belief Tracker} & 100\% & - & 0.865 & - \\
\hline
\emph{KVRN} & 100\% & 0.134 & - & N/A \\
\hline
\end{tabular}
\label{tab:mainre:camrest}
\end{table*}

\begin{table*}[t]
\caption{RQ1\&RQ2\&RQ4: Performance on Stanford In-Car Personal Assistant Corpus. Best performance is marked bold for each supervision proportion}
\centering
\begin{tabular}{c|c|c|c|c}
\hline
            &\textbf{Supervision Proportion} & \textbf{BLEU} &\textbf{Joint Goal Accuracy} &\textbf{Entity Match Rate}\\
\hline \hline
\multirow{3}{*}{\emph{SEDST}} &0\%	&\textbf{0.202} &\textbf{0.635} &\textbf{0.642}\\
                                     &25\%  &\textbf{0.192} &\textbf{0.758} &\textbf{0.813}\\
                                     &50\%  &\textbf{0.195} &\textbf{0.796} &\textbf{0.833}\\
\hline
\multirow{3}{*}{\emph{SEDST$\backslash$PR}}  & 0\%	 &0.193 &0.622 &0.564\\
                               & 25\% &0.180 &0.726 &0.770\\
							   & 50\% &0.178 &\textbf{0.796} &0.812\\ 
\hline                           
\multirow{3}{*}{\emph{SEDST(without unlabeled data)}}
									&0\%    & - & - & - \\
									&25\% 	&0.102 &0.727 &0.751\\
                                    &50\%	&0.156 &0.772 &0.773\\
\hline
\emph{SEDST(fully supervised)} & 100\% & \textbf{0.193} & \textbf{0.829} & \textbf{0.845} \\
\hline
\emph{NDM} & 100\% & 0.186 & 0.750 & 0.716 \\
\hline
\emph{Neural Belief Tracker} & 100\% & - & 0.756 & - \\
\hline
\emph{KVRN} & 100\% & 0.172 & - & 0.459 \\
\hline

\end{tabular}
\label{tab:mainre:kvret}
\end{table*}

\subsection{Baselines and comparisons}
\label{subsec:exp:baselines}

We list the methods and baselines below. We write \textbf{SEDST} for the overall process as described in~\S\ref{section4}, which includes posterior regularization. We write \textbf{SEDST$\backslash$PR} for the model that skips the posterior regularization process.

To assess the contribution of our proposed methods, our baselines include recent work on both task-oriented dialogue models and non-task-oriented dialogue models. We adopt the following baselines for task-oriented dialogue generation under fully supervision.
\begin{itemize}
			 \item \textbf{NDM}: Network based Dialogue Models \cite{wen-EtAl:2017:EACLlong} with a CNN-RNN dialogue state tracker
             \item \textbf{NBT}: Neural Belief Tracker \cite{mrkvsic2016neural} with a CNN feature extractor.
             \item \textbf{KVRN}: Key-value retrieval dialogue model \cite{eric2017key}, which does not adopt a dialogue state tracker but directly retrieves an entry from a key-value structured knowledge base with attention mechanism, and decodes a special token in <subject, relation, object> form during response generation.
\end{itemize}

We utilize the following representative baselines for non-task-oriented dialogue generation:
\begin{itemize}
  \item \textbf{SEQ2SEQ}: sequence-to-sequence model, also known as recurrent encoder-decoder model \cite{shang,vinyals2015neural}.
  \item \textbf{HRED}: hierarchical recurrent encoder-decoder model.\cite{Sordoni2015hre}
  \item \textbf{VHRED}: latent variable hierarchical recurrent encoder-decoder model. \cite{hvred}.
  \item \textbf{HVMN}: hierarchical variational memory network. ~\cite{chen18www}
  \item \textbf{DAWnet}: deep and wide neural network for dialogue generation. It first generates keywords that deepen or widen topics before response generation. Ground truth of keywords are obtained with rules and unsupervised methods~\cite{dawnet} 
\end{itemize}

\subsection{Evaluation metrics}
\label{subsec:exp:eval}
\paratitle{Task-oriented dialogue evaluation}

To assess the language quality and the state tracking ability for the task-oriented dialogue generation, we employ \textbf{BLEU}~\cite{Papineni2002BLEU}, a word-overlapping based metric for language quality evaluation, to measure performance. 

To measure the state tracker performance, we employ \textbf{Joint Goal Accuracy}~\cite{mrkvsic2016neural} as our turn-level evaluation metric. Joint goal accuracy calculates the proportion of the dialogue turns where all the constraints are captured correctly, excluding those where the user merely thanks without extra information. 

However, this metric is not applicable for task-oriented dialogue systems without a separate state tracker, such as KVRN. Thus we employ \textbf{Entity Match Rate}~\cite{wen-EtAl:2017:EACLlong}, as an evaluation metric. Entity Match Rate calculates the proportion of the dialogues where all the constraints are correctly identified when the last placeholder (e.g., poi\_SLOT) appears. We skip dialogues without a single placeholder in ground-truth responses; we consider dialogues as failures if no placeholder is decoded for remaining dialogues.

\paratitle{Non-task oriented dialogue evaluation} 

Evaluating dialogue systems in such a large corpus is not a trivial task.  \citet{Liu2016How} showed that word-overlap automatic metrics like \textbf{BLEU}~\cite{Papineni2002BLEU} or \textbf{ROUGE}~\cite{rouge2004} are not well correlated with human evaluations regarding response quality. To evaluate the semantic relevance between the candidate response and target response, we employ three embedding-based topic similarity metrics proposed by \citet{Liu2016How}: \textbf{Embedding Average}, \textbf{Embedding Extrema} and \textbf{Embedding Greedy}~\cite{mitchell2008,forgues2014bootstrapping,rus2012comparison}. We employ publicly available \emph{word2vec}\footnote{https://code.google.com/archive/p/word2vec/} to train word embeddings for evaluation. We train English word embeddings on Google News Corpus. For Chinese, the word embeddings are trained on Chinese Giga-word corpus version 5~\cite{graff2005chinese}, segmented by \emph{zpar}\footnote{https://github.com/SUTDNLP/ZPar} ~\cite{zhang2011syntactic}.

\subsection{Experimental settings}
\label{subsec:exp:training}

On Cambridge Restaurant corpus and Stanford In-Car Personal Assistant corpus, we trained the model with Adam~\cite{Kingma2015Adam} optimizer with a learning rate of 0.003 with early stopping. The batch-size was set to 32, and the size of the word embedding was set to 50. We used a single-layer GRU with 50 hidden units. The trade-off factor $\lambda$ was set to 0.1. For non-task-oriented dialogue models on Ubuntu Dialogue Corpus and JD.com Customer Service Corpus, we trained the model with Adam~\cite{Kingma2015Adam} optimizer, under a learning rate of 0.0005 with early stopping. The batch size was set to 24 and the size of the word embedding was set to 300. We used a single-layer GRU with 500 hidden units for these models. The trade-off factor $\lambda$ was set to 0.1 at the beginning, and uniformly decrease to 0.001 within the first epoch. We loaded pretrained fastText~\cite{bojanowski2017enriching} word vectors for all the models, and the vocabulary size on these datasets was limited to 800, 1400, 20000, and 20000, respectively. 

For response generation, we applied beam search decoding with a beam size of 5. As for dialogue state decoding in semi-supervised or fully-supervised scenario, we performed max-sampling on $\mathbf{p}_i$ to generate word sequences in state tracker with a special token indicating sequence termination. However, in unsupervised scenario, it is impossible for model to generate end-of-sequence tokens. Thus, our state decoder generated a sequence by a fixed time step of $T_s$ during decoding in unsupervised setting. In our experiments, $T_s$ was set to 5 on Ubuntu Technical Dialogue Corpus and 8 for all other corpora. We prevent generation of repeated words in the state span during sampling. During the evaluation of unsupervised dialogue state trackers on task-oriented corpora, we calculated the intersection of the output of state decoders and all possible slot values provided separately in the corpus. This setup is same as baselines where dialogue states are limited to a fix-sized set.

\section{Experimental Results}
\label{section6}

In \S\ref{sec61}, we compare our methods to baselines for task-oriented dialogue generation; in \S\ref{sec62} we examine the performance of comparisons for non-task-oriented dialogue generation; \S\ref{sec63} examines the effect of posterior regularization. We discuss the explainability and representativeness of dialogue state tracking in \S\ref{sec64}.

\subsection{Task-oriented dialogue systems}
\label{sec61}
To start, we address research question \textbf{RQ1} for task-oriented dialogue systems. Table~\ref{tab:mainre:camrest} and Table~\ref{tab:mainre:kvret} list the performance of all methods on two task-oriented dialogue corpora respectively. For all two datasets, under fully supervised settings, SEDST outperforms other baselines and achieves state-of-the-art performance. On Cambridge corpus, SEDST with fully supervision achieves a $2.09\%$, $4.45\%$, and $5.87\%$ over NDM in terms of BLEU, joint goal accuracy, and entity math rate, respectively; whereas on Stanford corpus, it achieves a $3.76\%$, $10.5\%$, and $18.0\%$, respectively.

To address research question \textbf{RQ2}, we analyze the performance of our models in semi-supervised settings where only part of labeled data is available. We find that SEDST outperforms SEDST trained without unlabeled data in both corpora on all corpora. 
In details, when only 25\% of data is labeled, in terms of joint goal accuracy, SEDST offers a 4.0\% and 5.1\% increase on two corpora respectively; whereas it gives 3.1\% and 6.2\% increase in terms of entity match rate. When only 50\% of data is labeled, its increase become 4.9\%, 2.8\% and 2.4\%, 6.0\% respectively. It verifies that our model is capable to utilize unlabeled data well for training dialogue state trackers. Moreover, we find our model outperforms state-of-the-art baselines on two corpora for both state tracking ability and language quality when only 50\% of labeled data is available. 
We notice that SEDST provides fairly satisfying results even if the state tracker is trained in a complete unsupervised manner. In terms of joint goal accuracy, SEDST achieves 68.4\% and 63.5\% on two corpora respectively; whereas entity match rate performance becomes 64.9\% and 64.2\%. 
\vspace*{-.\baselineskip}  

\subsection{Non-task oriented dialogue systems}
\label{sec62}

Next, we turn to \textbf{RQ3}. From table~\ref{tab:metricub} and table~\ref{tab:metricjd}, We find $SEDST$ produces quite competitive results against state-of-the-art non-task-oriented dialogue models. Quite notably, on all corpora SEDST outperforms VHRED and HVMN, which uses continuous latent variables to maintain dialogue states. In addition, our state tracker produces explicit dialogue states in state trackers. We notice that the entities mentioned between context and generated responses are highly relevant, which effectively tackles a key challenge in neural dialogue generation. Table~\ref{case:nontask} provides examples of state tracker outputs and generated responses.
\vspace*{-.5\baselineskip}
\subsection{Effect of posterior regularization}
\label{sec63}

Turning to \textbf{RQ4}, shown in Table~\ref{tab:mainre:camrest} and Table~\ref{tab:mainre:kvret}, we find that our posterior regularization improves overall state tracking performance from the comparison between SEDST and SEDST$\backslash$PR. On Cambridge corpus, when only 25\% labeled data is available, SEDST offers a 1.3\% and 1.0\% increase over SEDST$\backslash$PR in terms of joint goal accuracy and entity match rate respectively; while it gives a 3.4\%, 2.6\% increase when 50\% labeled data is available. For Stanford corpus, we see a similar picture. It also improves response quality, as shown in Table~\ref{tab:metricub} and Table~\ref{tab:metricjd}.

Here we discuss the effect of posterior regularization.
In terms of state tracking in task-oriented dialogues, we train the posterior network with more informative input by including current turn response $R_t$; then we optimize the prior network by minimizing the distance of posterior and prior distributions (Eq.~\ref{eq:klpq}). Accordingly, SEDST performs better than SEDST$\backslash$PR when fewer labeled data is available.
Posterior regularization also helps the response generation with unlabeled dialogues states.
Although the prior network can explore a generation strategy of $S_t$, $S_t$ is also regularized towards the compressed representation of context and responses learned by the posterior network. 
The representation which we regularize towards is a probability distribution over the vocabulary space learned by auto-encoders. 
We notice a state-of-the-art work, DAWnet~\cite{dawnet}, extracts keywords with rules and unsupervised methods from unlabeled data. Key word prediction is then trained with supervised methods with maximum likely hood estimation objective in the neural network. 
However, posterior regularization is a generalization of the learning strategy of DAWnet~\cite{dawnet}. When the $i$-th word over the vocabulary is treated as the ground truth of the keyword, the objective of maximum likelihood estimation is equivalent to minimizing the KL-divergence between the prior distribution $\mathbf{p}$ and an one-hot distribution $\mathbf{\hat{q}}$, since we have:
\vspace*{-.5\baselineskip}
\begin{equation}
log(p_i) = -KL(\mathbf{\hat{q}}||\mathbf{p}).
\end{equation}

\noindent where $\mathbf{\hat{q}}$ is a multinomial distribution with $\mathbf{\hat{q}_{i}}=1$ at its $i$-th coordinate. However, unlike DAWnet, posterior regularization provides probability distributions over the whole vocabulary space as learning signals, which are more informative than a handful of keywords and immune to extraction bias from rules. Moreover, our model is fully end-to-end trainable.

We notice that the text spans in DAWnet are trained on \emph{predicted keywords}~\cite{dawnet}, which are defined as the keywords that appear in ground truth responses but not appear in contexts. As an empirical study of the contribution of $S_t$, in Table~\ref{tab:kw} we present the proportion of generated \emph{predicted keywords} that exist in ground truth responses. In contrast with DAWnet, SEDST gives an obviously larger proportion of correct \emph{predicted keywords} for both corpora.

\begin{table}[!t]
\caption{RQ3: Embedding-based evaluation in Ubuntu Technical Corpus. Emb. is an abbreviation for Embedding}
\centering
  \begin{tabular}{c|c|c|c}
  \hline
    \textbf{Model} & \textbf{Emb. Average} & \textbf{Emb. Greedy} & \textbf{Emb. Extrema}\\
  \hline
  \hline
    SEQ2SEQ        & 0.216           & 0.169           & 0.126          \\ 
    HRED           & 0.542           & 0.412           & 0.319          \\ 
    VHRED          & 0.534           & 0.403           & 0.306         \\ 
    HVMN		   & 0.558           & 0.423           & 0.322             \\
    DAWnet	       & 0.530           & 0.390           &  0.333             \\
  \hline
    SEDST$\backslash$PR         & 0.586	    & 0.438		& 0.330         \\ 
    SEDST &  \textbf{0.609}       &\textbf{0.451}		&\textbf{0.337}
      \\
  \hline
  \end{tabular}
  \label{tab:metricub}
  \vspace{-1.0em}
\end{table}
\begin{table}[!t]
\caption{RQ3: Embedding-based evaluation in JD Corpus. Emb. is an abbreviation for Embedding}
\centering
  \begin{tabular}{c|c|c|c}
  \hline
    \textbf{Model} & \textbf{Emb. Average} & \textbf{Emb. Greedy} & \textbf{Emb. Extrema}\\
  \hline
  \hline
    SEQ2SEQ        & 0.425            & 0.479         &  0.264          \\ 
    HRED           & 0.549        & 0.587        & \textbf{0.406}          \\ 
    VHRED          & 0.576       & 0.593         & 0.392         \\ 
    HVMN		   & 0.564       & 0.596         &  0.405          \\
    DAWnet	       & 0.579              &    0.574      &     0.375     \\
  \hline
    SEDST$\backslash$PR         & 0.575		    & 0.602		& 0.373         \\ 
    SEDST &  \textbf{0.585}   &  \textbf{0.607}		&  0.392
      \\
  \hline
  \end{tabular}
  \label{tab:metricjd}
  \vspace{-1.0em}
\end{table}

\subsection{Case study}
\label{sec64}

Finally, we address \textbf{RQ5}. In Table~\ref{case:task} and \ref{case:nontask} we provide example dialogues in task-oriented dialogue corpus and non-task-oriented dialogue corpus under unsupervised training. We see our model successfully captures informative words indicating dialogue states without any supervision. Especially in examples from the Ubuntu corpus, our state tracker generates explainable and representative words of dialogue process rather than generic words like ``the'' or ``is'', which co-occurs much more often in the whole corpus. 
Some important words are passed through dialogue turns, indicating that explicit tracking of dialogue states with copying mechanism is effective for tackling long-term dependency. The representative words are also addressed as \emph{topic continuity}~\cite{Lei2018} between turns.

\begin{table}[t]
\caption{RQ4: Proportion of generated \emph{predicted keywords}~\cite{dawnet} existing in ground truth responses.}
\centering
  \begin{tabular}{c|c|c}
  \hline
    \textbf{Model} & \textbf{Ubuntu Technical} & \textbf{Jd.com Customer Service} \\
  \hline
  \hline
    DAWnet       & 5.5\%        & 32.6\% \\    \hline
    SEDST        & 14.7\%         & 40.6\% \\
  \hline
  \end{tabular}
  \label{tab:kw}
  \vspace{-1.0em}
\end{table}
\vspace*{-.3\baselineskip}

\begin{table*}[!t]
\caption{Examples task-oriented dialogues in unsupervised training on Standford In-Car Assistant Corpus and Cambridge Restaurant Corpus}

  \centering
    \tiny
  \begin{tabular}{m{4cm}|m{6cm}|m{6cm}}
  \hline
  \textbf{Contexts} &\textbf{Generated} &\textbf{Ground Truth}\\
  \hline
    \multirow{2}{4cm}{i am looking for a german restaurant in the west part of town .} 
  &\textbf{State}: german west & \textbf{State}: west german \\
  &\textbf{Generated}: there are no german restaurants in the west part of town . would you like to try something else ?
  & \textbf{Response}: there are no german restaurants in the west part of town . would you like to make another query ?
\\
  \hline
  \multirow{2}{4cm}{how about italian ?}
   &\textbf{State}: italian west
 & \textbf{State}: italian west \\
   &\textbf{Generated}: name\_SLOT is located in the west part of town .
   & \textbf{Response}: there are several italian restaurants in the west part of town . name\_SLOT is pricerange\_SLOT and name\_SLOT ispricerange\_SLOT . would you like the address of one of those ?
\\
  \hline
    \multirow{2}{4cm}{what is the phone number of one of them ?}
   &\textbf{State}: italian west & \textbf{State}: italian west \\
   &\textbf{Generated}:name\_SLOT is located at address\_SLOT and their phone number is phone\_SLOT .
   & \textbf{Response}: the phone number to name\_SLOT , which is pricerange\_SLOT priced , is phone\_SLOT . is there anything else i can assist you with today ?
  \\
  \hline
  
  \end{tabular}
  \label{case:task}
\end{table*}

\begin{table*}[!t]
\caption{Examples dialogues on Ubuntu Technical corpus. The words that co-occur between generated dialogue state spans and responses are marked in italic}

  \centering
    \tiny
  \begin{tabular}{m{4cm}|m{6cm}|m{6cm}}
    \hline
  \textbf{Contexts} &\textbf{Generated} &\textbf{Ground Truth}\\
   \hline
      \multirow{4}{4cm}{I just realised nothing can access my mysql database anymore , it says cannot connect , why would this be ? I did set it to be remote}
   &\textbf{State}: remote would anymore mysql access
 & \textbf{State}: - \\
   &\textbf{Response}: you can use the \emph{mysql} client to connect to the server
   & \textbf{Response}: is it running ?
   \\
  & & \\
  & & \\
   \hline
   \multirow{2}{4cm}{is it running ?}
   &\textbf{State}: running I says mysql access & \textbf{State}: - \\
   &\textbf{Response}: yes , \emph{I} am \emph{running} it now
   & \textbf{Response}: How do \emph{I} check ? \emph{I} was able to type \emph{mysql} commands
   \\
   \hline
   \multirow{2}{4cm}{How do I check ? I was able to type mysql commands}
   &\textbf{State}: commands mysql I running How & \textbf{State}: - \\
   &\textbf{Response}: you can use sudo
   & \textbf{Response}: thats not mean your server was \emph{running} check with ps or top or others tool
   \\
   \hline
   \multirow{2}{4cm}{thats not mean your server was running check with ps or top or others tool}
   &\textbf{State}: mysql top tool server check& \textbf{State}: - \\
   &\textbf{Response}: thanks
   & \textbf{Response}: How do I \emph{check} if it 's running ? Ah , k 
   \\
   \hline
   \multirow{2}{4cm}{How do I check if it 's running ? Ah , k }
   &\textbf{State}:mysql check ps server thats & \textbf{State}: - \\
   &\textbf{Response}: \emph{ps} aux | grep \emph{mysql}
   & \textbf{Response}:\emph{ps} aux | grep mysqld
   \\
   \hline

   \hline
  \end{tabular}
  \label{case:nontask}
\end{table*}

\section{Conclusion}
\label{section7}
We have addressed the task of explicit dialogue state tracking for both task-oriented and non-task-oriented dialogue generation. We have identified challenges of dialogue state tracking: expensive nature of state labeling and weak explainability. To tackle these challenges, we have proposed a semi-supervised neural dialogue generation framework, named SEDST. Along with SEDST, we have proposed an explicit dialogue state tracking model, CopyFlowNet, including an implicit copyNet mechanism for representing explicit word sequences and a posterior regularization strategy for semi-supervised learning procedure.
In our experiments, we have presented a detailed comparison and deep analysis of the performance of SEDST, and have verified its effectiveness in terms of the dialogue generation quality and the dialogue state tracking accuracy on both non-tasked oriented dialogue corpora and task-oriented dialogue corpora.

As to future work, we plan to transfer our model to other tasks such as reading comprehension, filtering, and summarization~\cite{Rajpurkar2016SQuAD10,ren2016time,li2018deep}. 
Also, we would like to apply reinforcement learning to improve the performance of dialogue generation.

\bibliographystyle{abbrvnatnourl}
\bibliography{main}

\end{document}